\documentclass[10pt,journal]{IEEEtran}

\usepackage[T1]{fontenc}
\usepackage{amsmath,amssymb,amsfonts}
\usepackage{algorithm}
\usepackage{algorithmic}
\usepackage{graphicx}
\usepackage{booktabs}
\usepackage{multirow}
\usepackage{placeins}
\usepackage{cite}
\usepackage{url}
\usepackage[hidelinks]{hyperref}

\title{Gradient-Energy Guided Block-Wise Perturbations for Sharpness-Aware Minimization}

\author{Zhen Huang, Jiaxin Deng, and Junbiao Pang%
\thanks{Z. Huang, J. Deng, and J. Pang are with the Faculty of Information
Technology, Beijing University of Technology, Beijing, China (e-mail:
huangzhen@emails.bjut.edu.cn; dengjiaxin@emails.bjut.edu.cn;
junbiao\_pang@bjut.edu.cn).}%
\thanks{Corresponding author: Junbiao Pang.}}

\begin{document}

\maketitle

\begin{abstract}
Sharpness-Aware Minimization (SAM) improves generalization by minimizing the worst-case loss in a local parameter neighborhood. Standard SAM implicitly allocates its global perturbation budget across parameter blocks according to instantaneous minibatch gradient norms. Such an allocation can be noisy and may not reflect the sensitivity that blocks accumulate throughout training. We propose Gradient-Energy Adaptive Radius SAM (GEAR-SAM), which maintains an exponential moving average (EMA) of squared block gradients as a lightweight, curvature-related sensitivity signal and allocates the fixed SAM budget through a closed-form constrained optimization. GEAR-SAM preserves the global SAM radius, requires no Hessian-vector products or explicit Fisher estimation, and adds only scalar state beyond SAM. Experiments on image classification, transfer learning, noisy-label learning, and partition studies demonstrate improved generalization and robustness across architectures and tasks. More broadly, GEAR-SAM provides a dynamic view of sharpness-aware optimization: a fixed perturbation budget should be redistributed as the sensitivity of functional network blocks evolves during training.
\end{abstract}

\section{Introduction}

Deep neural networks often generalize better when optimization converges to flatter minima \cite{keskar2016large,jiang2019fantastic,chaudhari2019entropy}. Sharpness-Aware Minimization (SAM) explicitly encourages such solutions by minimizing the worst-case loss in a local neighborhood of the current parameters \cite{foret2020sharpness}. Due to its simplicity and effectiveness, SAM has become a widely used training strategy and has inspired many follow-up methods, including ASAM \cite{kwon2021asam}, GSAM \cite{zhuang2022surrogate}, Fisher SAM \cite{kim2022fisher}, F-SAM \cite{li2024friendly}, and BSAM \cite{deng2026bilateral}.

Despite its success, standard SAM makes its perturbation decision from the current stochastic gradient. When the network is partitioned into functional blocks, this decision implicitly assigns a different portion of the global perturbation budget to each block, but the allocation is determined only by the current minibatch. This is a dynamic issue: the optimization roles and perturbation sensitivities of network blocks can evolve throughout training, whereas an instantaneous gradient can be dominated by minibatch noise or transient updates. Thus, a fixed global radius should not imply a fixed or purely instantaneous allocation across the network.

To construct a stable dynamic allocation, the signal should reflect more than a transient gradient magnitude. Directly using curvature information during training is difficult: Hessian matrices are infeasible to store for modern networks, and online Hessian-vector or Hessian-block estimation introduces additional computation. Fisher information provides a natural curvature-related alternative and has been used in geometry-aware SAM methods \cite{kim2022fisher}. Nevertheless, explicit Fisher estimation is still costly. Motivated by the connection between squared gradients, diagonal empirical Fisher information, and expected Hessian information under standard likelihood assumptions \cite{kunstner2019limitations,hwang2024fadam}, we use an online second-moment statistic as a lightweight curvature-related signal.

In this work, we revisit SAM as a dynamic block-wise budget-allocation method. We show that standard SAM induces a different effective radius for each block, but this implicit allocation depends only on instantaneous minibatch gradient norms. GEAR-SAM instead uses an exponential moving average (EMA) of squared block gradients as a persistent sensitivity score and allocates the fixed SAM perturbation budget according to this historical signal. The resulting perturbation preserves the global SAM budget while assigning larger radii to blocks that have remained sensitive over the recent optimization history.

Our contributions are summarized as follows:
\begin{itemize}
    \item We analyze SAM from a second-order and dynamic block-wise perspective, showing that its practical perturbation is first-order optimal but ignores curvature-related block sensitivity and temporal persistence.

    \item We propose GEAR-SAM, which uses EMA-based historical gradient energy as a persistent Fisher-related second-moment signal and derives a closed-form block-wise radius allocation under a fixed SAM budget.

    \item We conduct experiments on image classification, transfer learning, label-noise robustness, and partition and parameter studies, demonstrating the effectiveness and stability of the proposed allocation strategy.
\end{itemize}

\section{Related Work}

\subsection{Sharpness-Aware Minimization}

SAM improves generalization by optimizing the loss at adversarially perturbed parameters within a local neighborhood \cite{foret2020sharpness}. The method is simple and effective, but its practical perturbation is computed using a first-order approximation of the inner maximization. A large body of work has therefore studied how to improve SAM from different perspectives. ASAM addresses the scale dependency of sharpness \cite{kwon2021asam}. GSAM introduces a surrogate gap to better guide flat-minima search \cite{zhuang2022surrogate}. F-SAM analyzes the role of stochastic-gradient components in SAM perturbations \cite{li2024friendly}. BSAM introduces bilateral sharpness by considering both max-sharpness and min-sharpness \cite{deng2026bilateral}. Other methods improve the efficiency of SAM through stochastic perturbation, perturbation reuse, sample selection, or parallelization \cite{du2021efficient,liu2022towards,jiang2023adaptive,ni2022k,xie2024sampa}. These methods mainly modify the sharpness objective, perturbation direction, or computational procedure. In contrast, GEAR-SAM focuses on how a fixed perturbation budget is allocated across parameter blocks.

\subsection{Structure-aware SAM}

Although SAM is usually formulated over the full parameter vector, recent work shows that its behavior depends strongly on network structure. Perturbing normalization layers can retain much of the benefit of full-parameter SAM \cite{mueller2023normalization}, and the role of normalization has been studied theoretically and empirically \cite{dai2023crucial}. Effective SAM updates may require layer-wise perturbation scaling \cite{haas2024mu}, while sparse or selective perturbations can further improve SAM \cite{mi2022make,zhong2022improving,lee2024layer,cheng2026sparse}. These findings indicate that treating the network as a single undifferentiated parameter vector can overlook useful structure.

These studies motivate structure-aware perturbation design. Most existing methods either select a subset of parameters or modify scaling at the layer level. GEAR-SAM instead partitions the network into functional blocks and continuously allocates a fixed global perturbation budget among them according to historical block sensitivity. It therefore changes neither the perturbation geometry nor the set of perturbed parameters.

\subsection{Curvature Proxies for Hessian}

Sharpness-aware training is closely related to local curvature. Hessian-based quantities, such as the largest eigenvalue or trace, are widely used to characterize flatness and generalization \cite{keskar2016large,jiang2019fantastic,jastrzkebski2017three,chaudhari2019entropy}. Recent studies also connect SAM with stability, first-order flatness, and explicit curvature regularization \cite{long2024sharpness,zhang2023gradient,wu2024cr}. However, explicit Hessian computation is expensive for modern neural networks.

Fisher information provides a practical curvature-related alternative. Fisher SAM uses Fisher geometry to define SAM neighborhoods \cite{kim2022fisher}, while Fisher Mask SAM uses Fisher information to estimate parameter importance \cite{zhong2022improving}. Nevertheless, explicit Fisher estimation is still costly, and empirical Fisher should not be treated as an exact Hessian estimator \cite{kunstner2019limitations}. Adam-style squared-gradient statistics provide an efficient online second-moment signal \cite{kingma2014adam} and have been connected to diagonal empirical Fisher information \cite{hwang2024fadam}. Motivated by this connection, GEAR-SAM uses block-wise historical gradient energy as a lightweight curvature-related proxy for adaptive radius allocation.

\section{Method}

\subsection{Motivation}

SAM solves the following local min-max problem:
\begin{equation}
    \min_{\mathbf{w}}
    \max_{\|\boldsymbol{\epsilon}\|_2\le\rho}
    L(\mathbf{w}+\boldsymbol{\epsilon}).
\end{equation}
At iteration $t$, let
\begin{equation}
    \mathbf{g}_t=\nabla_{\mathbf{w}}L_{\mathcal{B}_t}(\mathbf{w}_t),
    \quad
    \mathbf{H}_t=\nabla_{\mathbf{w}}^2L_{\mathcal{B}_t}(\mathbf{w}_t).
\end{equation}
The practical perturbation used by SAM is derived from the first-order approximation
\begin{equation}
    L_{\mathcal{B}_t}(\mathbf{w}_t+\boldsymbol{\epsilon})
    \approx
    L_{\mathcal{B}_t}(\mathbf{w}_t)
    +
    \mathbf{g}_t^\top\boldsymbol{\epsilon}.
\end{equation}
Solving the corresponding inner maximization gives
\begin{equation}
    \boldsymbol{\epsilon}^{\mathrm{SAM}}_t
    =
    \rho
    \frac{\mathbf{g}_t}{\|\mathbf{g}_t\|_2}.
    \label{eq:first_order_sam}
\end{equation}
for $\mathbf{g}_t\neq\mathbf{0}$; if the full gradient is zero, we define the
first-order perturbation as zero.
Thus, Eq.~\eqref{eq:first_order_sam} is optimal for the linearized sharpness
objective. This follows directly from the Cauchy--Schwarz inequality:
$\mathbf{g}_t^\top\boldsymbol{\epsilon}\le
\|\mathbf{g}_t\|_2\|\boldsymbol{\epsilon}\|_2\le
\rho\|\mathbf{g}_t\|_2$, with equality when
$\boldsymbol{\epsilon}$ is parallel to $\mathbf{g}_t$.

However, the local sharpness objective also contains curvature. A second-order approximation gives
\begin{equation}
    L_{\mathcal{B}_t}(\mathbf{w}_t+\boldsymbol{\epsilon})
    \approx
    L_{\mathcal{B}_t}(\mathbf{w}_t)
    +
    \mathbf{g}_t^\top\boldsymbol{\epsilon}
    +
    \underbrace{\frac{1}{2}
    \boldsymbol{\epsilon}^{\top}\mathbf{H}_t\boldsymbol{\epsilon}}_{Q_t(\boldsymbol{\epsilon})}.
    \label{eq:second_order_expansion}
\end{equation}
Let $Q_t(\boldsymbol{\epsilon})$ denote the quadratic terms in
Eq.~\eqref{eq:second_order_expansion}. The maximization of $Q_t$ over the
norm ball is a classical trust-region problem \cite{more1983computing}. When
the associated optimality system is nonsingular, its solution has the form
\begin{equation}
    \boldsymbol{\epsilon}_t^\star
    =
    (2\lambda_t\mathbf{I}-\mathbf{H}_t)^{-1}\mathbf{g}_t,
    \label{eq:second_order_direction}
\end{equation}
where $\lambda_t$ is chosen to satisfy the trust-region conditions. The
complete conditions, including the singular hard case, are given in the
appendix. Equation~\eqref{eq:second_order_direction} shows that the desirable
perturbation is shaped by both gradient $\mathbf{g}_t$ and curvature $\mathbf{H}_t$. Thus, SAM is
first-order optimal, but it is generally not optimal for the second-order
sharpness objective. 

\subsection{Block-Wise View of SAM}

Computing the Hessian information in Eq.~\eqref{eq:second_order_direction}
online is costly. We therefore study how to allocate the fixed SAM radius across
parameter blocks while preserving the current gradient direction within each
block. Let the parameters be divided into $B$ disjoint blocks:
\begin{equation}
    \mathbf{w}
    =
    (\mathbf{w}_1,\mathbf{w}_2,\ldots,\mathbf{w}_B),
\end{equation}
with the corresponding gradient decomposition
\begin{equation}
    \mathbf{g}_t
    =
    (\mathbf{g}_{1,t},\mathbf{g}_{2,t},\ldots,\mathbf{g}_{B,t}).
\end{equation}
\noindent\textbf{Architecture-aware block partition.}
We use a coarse, architecture-aware partition rather than treating every
parameter tensor as an independent group. A block is a self-contained
architecture-level computation unit, or a contiguous group of such units that
operates at the same representation scale. Thus, a coarse block can contain
multiple residual units; the input stem and final classifier are separate
blocks. Every trainable parameter belongs to exactly one block.

All trainable affine normalization parameters belonging to a computation unit
are assigned to the same block as its weight layers; non-trainable
normalization buffers are not perturbed. This treatment is important because
normalization parameters are not interchangeable with arbitrary small
parameter subsets. In particular, perturbing only affine normalization
parameters can retain, and sometimes exceed, the benefit of full-parameter SAM
\cite{mueller2023normalization}. Assigning these parameters to a separate
radius would split the scale-modulating normalization operation from the
weights whose activations it normalizes. We therefore use the aggregate
gradient energy of the complete functional block as its allocation signal.
This is an architecture-aware design principle, not a claim that a single
partition granularity is universally optimal; alternative block granularities
are evaluated in Table~\ref{tab:partition_strategies}.

From Eq.~\eqref{eq:first_order_sam}, the perturbation applied to block $b$ is
\begin{equation}
    \boldsymbol{\epsilon}^{\mathrm{SAM}}_{b,t}
    =
    \rho
    \frac{\mathbf{g}_{b,t}}
    {\sqrt{\sum_{j=1}^{B}\|\mathbf{g}_{j,t}\|_2^2}}.
\end{equation}
This can be written as a block-wise radius times a normalized block direction:
\begin{equation}
    \boldsymbol{\epsilon}^{\mathrm{SAM}}_{b,t}
    =
    r^{\mathrm{SAM}}_{b,t}
    \frac{\mathbf{g}_{b,t}}{\|\mathbf{g}_{b,t}\|_2},
    \quad
    r^{\mathrm{SAM}}_{b,t}
    =
    \rho
    \frac{\|\mathbf{g}_{b,t}\|_2}
    {\sqrt{\sum_{j=1}^{B}\|\mathbf{g}_{j,t}\|_2^2}}.
    \label{eq:sam_block_radius}
\end{equation}
Equation~\eqref{eq:sam_block_radius} reveals that SAM already performs
block-wise radius allocation. Its allocation score is the instantaneous
minibatch gradient norm $\|\mathbf{g}_{b,t}\|_2$. This allocation is
consistent with the first-order inner maximization, but it does not encode the
curvature term in Eq.~\eqref{eq:second_order_expansion}.

\noindent\textbf{Observation 1 (Implicit SAM allocation).}
For any partition of parameters into blocks and any nonzero full gradient,
the SAM perturbation in Eq.~\eqref{eq:first_order_sam} is equivalent to a
block-wise perturbation whose block radius is given by
Eq.~\eqref{eq:sam_block_radius}. Moreover, these block radii preserve the
global SAM budget, i.e.,
$\sum_{b=1}^{B}(r^{\mathrm{SAM}}_{b,t})^2=\rho^2$.
The derivation is given in the appendix. This observation makes explicit that
SAM is already an implicit allocation method, but the allocation signal is
only the instantaneous gradient norm. If the full gradient is zero, SAM uses
zero perturbation under our convention. When only a block gradient is zero,
its SAM radius and perturbation are defined as zero.

\begin{figure*}[t]
    \centering
    \includegraphics[width=\textwidth]{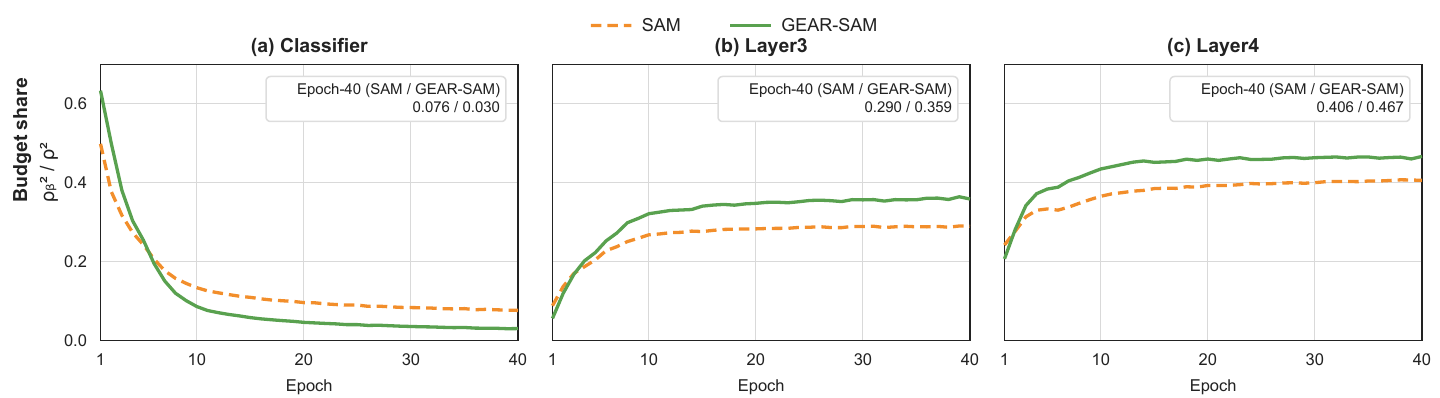}
    \caption{Early block-wise perturbation dynamics of SAM and GEAR-SAM on
    ResNet-18 trained on CIFAR-100. The plots show the first 40 epochs for the
    classifier and the two deeper feature blocks. Each panel reports the
    normalized perturbation-budget share $\rho_{b,t}^{2}/\rho^{2}$ at epoch 40.}
    \label{fig:block_dynamics}
\end{figure*}

Figure~\ref{fig:block_dynamics} illustrates this dynamic allocation on
ResNet-18 during the first 40 epochs, when the allocation changes most
rapidly. Although both methods use the same global perturbation budget at every
iteration, GEAR-SAM more quickly reduces the share assigned to the classifier
and assigns larger shares to Layer3 and Layer4. The complete trajectories are
reported in the appendix. This early redistribution motivates an allocation
signal that summarizes recent optimization history rather than reacting only to
the current minibatch.

To see the second-order information missing from the instantaneous allocation,
write the block perturbation as
\begin{equation}
    \boldsymbol{\epsilon}_{b,t}=r_{b,t}\mathbf{u}_{b,t},
    \quad
    \mathbf{u}_{b,t}=
    \frac{\mathbf{g}_{b,t}}{\|\mathbf{g}_{b,t}\|_2}.
\end{equation}
Here $\mathbf{u}_{b,t}$ is defined for blocks with nonzero gradient;
zero-gradient blocks contribute no first-order direction under our convention.
Let $\Delta L_t=L_{\mathcal{B}_t}(\mathbf{w}_t+\boldsymbol{\epsilon}_t)-
L_{\mathcal{B}_t}(\mathbf{w}_t)$ denote the perturbed minibatch-loss increase.
Applying a second-order Taylor expansion and, for interpretation, omitting
cross-block Hessian terms gives
\begin{equation}
\begin{aligned}
    \Delta L_t
    &\approx
    \sum_{b=1}^{B}
    r_{b,t}\|\mathbf{g}_{b,t}\|_2
    +
    \frac{1}{2}
    \sum_{b=1}^{B}
    r_{b,t}^{2}
    \mathbf{u}_{b,t}^{\top}
    \mathbf{H}_{bb,t}
    \mathbf{u}_{b,t}.
\end{aligned}
\label{eq:block_second_order}
\end{equation}
Therefore, an informative block-wise allocation should consider not only the
current gradient magnitude but also persistent block sensitivity related to
curvature.

\subsection{Gradient Energy as a Curvature-related Proxy}

The ideal block sensitivity in Eq.~\eqref{eq:block_second_order} would involve
Hessian-block information such as
$\mathbf{u}_{b,t}^{\top}\mathbf{H}_{bb,t}\mathbf{u}_{b,t}$, trace, or
spectral quantities. These quantities are too expensive to estimate at every
iteration. We instead use squared-gradient statistics as a lightweight online
proxy.

To distinguish per-example and minibatch quantities, let
$\mathbf{z}_t(x)=\nabla_{\mathbf{w}}\ell(\mathbf{w}_t;x)$
denote the gradient of one example $x$. Define its mean and
second-moment matrix,
\begin{equation}
    \boldsymbol{\mu}_t=\mathbb{E}_{x}[\mathbf{z}_t(x)],
    \qquad
    \mathbf{C}_t=\mathbb{E}_{x}
    [\mathbf{z}_t(x)\mathbf{z}_t(x)^{\top}].
    \label{eq:gradient_second_moment}
\end{equation}
For a negative log-likelihood, $\mathbf{C}_t$ equals the Fisher information
when the expectation is taken under the model distribution. Under the training
distribution, it is an empirical-Fisher-related gradient second moment and
should not be identified with the Hessian in general
\cite{kunstner2019limitations}. Under standard likelihood regularity and
model-matching assumptions, however, the Fisher equals the expected Hessian.
We use only this curvature-related connection, not an exact Hessian
equivalence.

Assume that the minibatch loss is an average over $m$ independently sampled
examples. For block $b$,
\begin{equation}
    \mathbf{g}_{b,t}
    =
    \frac{1}{m}\sum_{i=1}^{m}\mathbf{z}_{b,t}(x_i).
\end{equation}
Conditioned on $\mathbf{w}_t$, its squared norm satisfies the exact identity
\begin{equation}
\begin{aligned}
    \mathbb{E}_{\mathcal{B}_t}\|\mathbf{g}_{b,t}\|_2^2
    ={}&
    \frac{1}{m}\operatorname{Tr}(\mathbf{C}_{bb,t}) \\
    &+
    \left(1-\frac{1}{m}\right)
    \|\boldsymbol{\mu}_{b,t}\|_2^2.
\end{aligned}
\label{eq:minibatch_second_moment}
\end{equation}
Thus, the squared minibatch gradient is not an unbiased empirical-Fisher
trace estimator. Near a stationary region, where
$\|\boldsymbol{\mu}_{b,t}\|_2$ is small,
$m\,\mathbb{E}\|\mathbf{g}_{b,t}\|_2^2$ approaches the block-wise
second-moment trace; the common factor $m$ cancels after block normalization.
Away from stationarity, the statistic also retains a mean-gradient
contribution, making it a mixed first- and second-moment sensitivity signal.

For block $b$, we define the instantaneous gradient energy as
\begin{equation}
    e_{b,t}
    =
    \|\mathbf{g}_{b,t}\|_2^2,
\end{equation}
and maintain its exponential moving average:
\begin{equation}
    s_{b,t}
    =
    \beta s_{b,t-1}
    +
    (1-\beta)e_{b,t},
    \quad
    \beta\in[0,1).
    \label{eq:ema_energy}
\end{equation}
Because summation and EMA updates are linear, $s_{b,t}$ is equivalent to
summing Adam-style per-parameter second moments within block $b$
\cite{kingma2014adam,hwang2024fadam}. We deliberately retain the total
second-moment mass rather than divide by the block dimension: it is additive
over a partition and represents the aggregate sensitivity of a block under a
global $\ell_2$ perturbation budget.

\noindent\textbf{Lemma 1 (EMA estimation).}
Assume that the stochastic energies $e_{b,t}$ are locally stationary with
mean $\bar e_b$. With $s_{b,0}=0$, the EMA satisfies
\begin{equation}
    \mathbb{E}[s_{b,t}]
    =
    (1-\beta^t)\bar e_b.
    \label{eq:ema_expectation}
\end{equation}
The common bias factor $1-\beta^t$ cancels in the normalized allocation rule.
The appendix also shows that, under an independent-noise idealization, EMA
reduces the variance of the instantaneous energy. We therefore use
$s_{b,t}$ as a persistent aggregate sensitivity score, not as an exact Fisher
or Hessian estimator.

\subsection{Gradient-Energy Adaptive Radius Allocation}

Given the block sensitivity scores $\{s_{b,t}\}_{b=1}^{B}$, we allocate the global SAM radius $\rho$ across blocks. Let $r_{b,t}$ denote the radius assigned to block $b$. We require the same global budget as SAM:
\begin{equation}
    \sum_{b=1}^{B}r_{b,t}^{2}\le \rho^2,
    \quad
    r_{b,t}\ge 0.
\end{equation}
This leaves the question of how to transform sensitivity scores into radii.
We use a simple surrogate allocation principle: larger sensitivity should
receive a larger radius, the global budget should remain controlled, and the
rule should not introduce additional block-wise hyperparameters. These
requirements do not uniquely solve the second-order inner problem. We
instantiate them with a scale-invariant linear alignment surrogate:
\begin{equation}
    \max_{\{r_{b,t}\}_{b=1}^{B}}
    \sum_{b=1}^{B} s_{b,t} r_{b,t},
    \quad
    \mathrm{s.t.}
    \quad
    \sum_{b=1}^{B} r_{b,t}^2 \le \rho^2,
    \quad
    r_{b,t}\ge 0.
    \label{eq:allocation_problem}
\end{equation}
Equation~\eqref{eq:allocation_problem} is a budget-preserving surrogate rather
than an exact second-order solution. We align radii directly with the
second-moment mass $s_{b,t}$, which emphasizes blocks with persistently large
aggregate sensitivity, is invariant to a common rescaling of all scores, and
avoids an additional exponent hyperparameter.

\noindent\textbf{Proposition 1 (Closed-form allocation).}
If the nonnegative score vector is nonzero, the unique radius vector solving
Eq.~\eqref{eq:allocation_problem} is
\begin{equation}
    r_{b,t}^{\star}
    =
    \rho
    \frac{s_{b,t}}
    {\sqrt{\sum_{j=1}^{B}s_{j,t}^{2}}}.
    \label{eq:gear_radius}
\end{equation}
The proof is given in the appendix. We denote the resulting block radius by
$\rho_{b,t}=r_{b,t}^{\star}$.

The perturbation direction within each block is kept as the current gradient direction, giving
\begin{equation}
    \boldsymbol{\epsilon}_{b,t}
    =
    \rho_{b,t}
    \frac{\mathbf{g}_{b,t}}
    {\|\mathbf{g}_{b,t}\|_2}.
    \label{eq:gear_perturbation}
\end{equation}
The idealized analysis assumes nonzero block gradients. In implementation, we
use the stabilized direction
$\widetilde{\mathbf{u}}_{b,t}=\mathbf{g}_{b,t}/(\|\mathbf{g}_{b,t}\|_2+\delta)$
with a small $\delta>0$; a zero-gradient block therefore contributes no
perturbation. The complete perturbation is
\begin{equation}
    \boldsymbol{\epsilon}_{t}
    =
    (\boldsymbol{\epsilon}_{1,t},\boldsymbol{\epsilon}_{2,t},\ldots,\boldsymbol{\epsilon}_{B,t}).
\end{equation}

The allocated radii exactly exhaust the radius budget,
\begin{equation}
    \sum_{b=1}^{B}\rho_{b,t}^{2}=\rho^2.
\label{eq:budget_preservation}
\end{equation}
With exact unit directions, Eq.~\eqref{eq:budget_preservation} also gives
$\|\boldsymbol{\epsilon}_t\|_2=\rho$. With the stabilized implementation,
$\|\widetilde{\mathbf{u}}_{b,t}\|_2\le 1$, and hence
$\|\boldsymbol{\epsilon}_t\|_2\le\rho$. Therefore, GEAR-SAM never enlarges
the SAM neighborhood. Its effect comes from reallocating the same radius
budget according to historical curvature-related block sensitivity.

\subsection{Algorithm and Complexity}

After constructing $\boldsymbol{\epsilon}_t$, GEAR-SAM follows the standard SAM update rule. The perturbed gradient is computed as
\begin{equation}
    \mathbf{G}_t
    =
    \nabla_{\mathbf{w}}
    L_{\mathcal{B}_t}
    (\mathbf{w}_t+\boldsymbol{\epsilon}_t).
    \label{eq:perturbed_gradient}
\end{equation}
With SGD as the base optimizer, the parameter update is
\begin{equation}
    \mathbf{w}_{t+1}
    =
    \mathbf{w}_t
    -
    \eta \mathbf{G}_t.
    \label{eq:base_update}
\end{equation}
Other base optimizers can also be used in the same way.

\begin{algorithm}[t]
\caption{GEAR-SAM}
\label{alg:gear_sam}
\begin{algorithmic}[1]
\REQUIRE Training data, model parameters $\mathbf{w}$, learning rate $\eta$, SAM radius $\rho$, EMA factor $\beta$, stability constant $\delta$
\STATE Initialize $s_b=0$ for each block $b=1,\ldots,B$
\FOR{each iteration $t$}
    \STATE Sample a minibatch $\mathcal{B}_t$
    \STATE Compute block-wise gradients on $\mathcal{B}_t$
    \STATE Update block sensitivity scores according to Eq.~\eqref{eq:ema_energy}
    \IF{the score vector is nonzero}
        \STATE Allocate block radii according to Eq.~\eqref{eq:gear_radius}
    \ELSE
        \STATE Set $\rho_{b,t}=0$ for all blocks
    \ENDIF
    \STATE Construct stabilized block perturbations as described after Eq.~\eqref{eq:gear_perturbation}
    \STATE Compute the perturbed gradient according to Eq.~\eqref{eq:perturbed_gradient}
    \STATE Update the parameters according to Eq.~\eqref{eq:base_update}
\ENDFOR
\end{algorithmic}
\end{algorithm}

GEAR-SAM has the same number of forward--backward passes as SAM: one gradient
computation constructs the perturbation and a second computes the gradient at
the perturbed point. Computing the block energies requires reductions over the
model gradients, with $O(d)$ arithmetic for $d$ parameters, comparable to the
global-norm reduction already used by SAM. The persistent state consists of
only $B$ scalar EMA values, followed by $O(B)$ normalization. Thus GEAR-SAM
does not change the dominant forward--backward complexity and introduces only
lightweight reductions and scalar state beyond SAM.

Different from sparse or selective SAM methods, GEAR-SAM does not decide whether a layer should be perturbed. Instead, it continuously allocates radii across blocks under a fixed global budget. Different from Fisher-based SAM variants, it does not modify the perturbation geometry or require explicit Fisher matrix estimation. Thus, GEAR-SAM provides a lightweight plug-in allocation mechanism for sharpness-aware optimization.

\section{Experiments}

\subsection{Image Classification}

\subsubsection{Setup}

To evaluate GEAR-SAM's effectiveness, we conduct experiments on the CIFAR-10
and CIFAR-100 datasets across a variety of architectures, i.e., ResNet-18,
WideResNet-28-10, and PyramidNet-110. ResNet-18 and WideResNet-28-10 are
trained for 200 epochs, while PyramidNet-110 is trained for 300 epochs. For
ResNet-18 and WideResNet-28-10, the initial learning rate is set to 0.05 with
a cosine schedule, and the momentum and weight decay are set to 0.9 and
0.001, respectively. For PyramidNet-110, the initial learning rate is 0.1,
and the momentum and weight decay are set to 0.9 and 0.0005, respectively.
For CIFAR-10, $\rho$ is set to 0.1, while for CIFAR-100, it is set to 0.2.
The EMA factor $\beta$ is set to 0.9 for both datasets.

We apply the architecture-aware block partition described in the method
section. ResNet-18 is divided into the stem, four coarse residual blocks
(\texttt{layer1}--\texttt{layer4}), and the classifier. WideResNet-28-10 and
PyramidNet-110 are divided into the stem, three coarse representation blocks,
and the classifier. For GEAR-SAM, we conduct three independent runs with
different random seeds and report the average accuracy and standard
deviation.

We use vanilla SGD and SAM as baselines. To comprehensively evaluate the
performance, we also include ASAM, FisherSAM, F-SAM, SGD-SALR, Unified VaSSO,
SSAM-F, and SSAM-D for comparison. These methods are follow-up works of SAM
that aim to enhance generalization. For ASAM, FisherSAM, and F-SAM, we report
the results in F-SAM \cite{li2024friendly}. These three methods were run for
300 epochs on PyramidNet-110, and we denote them as ASAM$_{(300)}$,
FisherSAM$_{(300)}$, and F-SAM$_{(300)}$, respectively. In addition, we
compare several efficient SAM variants, including LookSAM, ESAM, and SAF.
For SGD-SALR, Unified VaSSO, SSAM-F, SSAM-D, LookSAM, ESAM, SAF, SGD, SAM,
and BSAM, we directly use the results reported in BSAM
\cite{deng2026bilateral}.

\subsubsection{Results}

Table~\ref{tab:cifar_results} reports the test accuracy on CIFAR-10 and
CIFAR-100. GEAR-SAM consistently improves upon SGD across all architectures
and datasets. On CIFAR-10, its largest gain over SGD is obtained with
PyramidNet-110, where accuracy increases by 0.71\%; it also improves standard
SAM by 0.09\% and 0.16\% on ResNet-18 and PyramidNet-110, respectively. On
CIFAR-100, the improvements are considerably larger. GEAR-SAM exceeds SGD by
1.85\%, 2.97\%, and 3.52\%, and exceeds SAM by 0.66\%, 0.82\%, and 0.84\% on
ResNet-18, WideResNet-28-10, and PyramidNet-110, respectively. This indicates
that the proposed allocation is effective across networks with different
depth and width configurations.

Compared with other SAM variants, GEAR-SAM achieves the best accuracy for
all three architectures on CIFAR-100, including both methods designed to
enhance generalization and efficient SAM variants. On CIFAR-10, it remains
competitive, while SSAM-D gives the highest accuracy on ResNet-18 and
WideResNet-28-10 and BSAM gives the highest accuracy on PyramidNet-110. The
smaller differences among methods on CIFAR-10 suggest that performance is
already close to saturation on this dataset. In contrast, the consistent
advantage on CIFAR-100 shows that historical block sensitivity becomes more
useful as the classification problem requires finer inter-class
discrimination. Overall, these results support the effectiveness of using a
stable curvature-related signal to guide perturbation allocation across
heterogeneous network blocks.

\begin{table}[!t]
\centering
\scriptsize
\caption{Test accuracy (\%) comparison of various networks on CIFAR-10 and CIFAR-100.}
\label{tab:cifar_results}
\setlength{\tabcolsep}{2pt}
\begin{tabular}{llcc}
\hline
Architecture & Method & CIFAR-10 & CIFAR-100 \\
\hline
\multirow{14}{*}{ResNet-18} & SGD & 96.18$\pm$0.09 & 79.89$\pm$0.38 \\
 & SAM & 96.74$\pm$0.05 & 81.08$\pm$0.27 \\
 & ASAM & 96.63$\pm$0.15 & 81.68$\pm$0.12 \\
 & FisherSAM & 96.72$\pm$0.03 & 80.99$\pm$0.13 \\
 & F-SAM & 96.75$\pm$0.09 & 81.29$\pm$0.12 \\
 & SGD-SALR & / & 81.73$\pm$0.10 \\
 & Unified VaSSO & 96.34$\pm$0.01 & 80.01$\pm$0.07 \\
 & SSAM-F & 96.81$\pm$0.02 & 81.24$\pm$0.21 \\
 & SSAM-D & \textbf{96.87$\pm$0.04} & 80.58$\pm$0.44 \\
 & LookSAM & 96.47$\pm$0.13 & 80.48$\pm$0.24 \\
 & ESAM & 96.56$\pm$0.08 & 80.41$\pm$0.10 \\
 & SAF & 96.37$\pm$0.02 & 80.06$\pm$0.05 \\
 & BSAM & 96.82$\pm$0.12 & 81.48$\pm$0.18 \\
 & GEAR-SAM & 96.83$\pm$0.10 & \textbf{81.74$\pm$0.08} \\
\hline
\multirow{14}{*}{WideResNet-28-10} & SGD & 96.93$\pm$0.05 & 82.56$\pm$0.27 \\
 & SAM & 97.52$\pm$0.05 & 84.71$\pm$0.21 \\
 & ASAM & 97.63$\pm$0.13 & 84.99$\pm$0.22 \\
 & FisherSAM & 97.46$\pm$0.18 & 84.91$\pm$0.07 \\
 & F-SAM & 97.53$\pm$0.11 & 85.16$\pm$0.07 \\
 & SGD-SALR & / & 83.12$\pm$0.09 \\
 & Unified VaSSO & 97.06$\pm$0.09 & 83.66$\pm$0.19 \\
 & SSAM-F & 97.71$\pm$0.23 & 85.16$\pm$0.96 \\
 & SSAM-D & \textbf{97.72$\pm$0.24} & 84.99$\pm$0.79 \\
 & LookSAM & 97.13$\pm$0.04 & 83.52$\pm$0.09 \\
 & ESAM & 97.29$\pm$0.11 & 84.51$\pm$0.01 \\
 & SAF & 97.08$\pm$0.15 & 83.81$\pm$0.04 \\
 & BSAM & 97.56$\pm$0.07 & 85.51$\pm$0.14 \\
 & GEAR-SAM & 97.50$\pm$0.05 & \textbf{85.53$\pm$0.04} \\
\hline
\multirow{11}{*}{PyramidNet-110} & SGD & 97.10$\pm$0.08 & 83.38$\pm$0.21 \\
 & SAM & 97.65$\pm$0.06 & 86.06$\pm$0.16 \\
 & ASAM$_{(300)}$ & 97.82$\pm$0.07 & 86.47$\pm$0.09 \\
 & FisherSAM$_{(300)}$ & 97.64$\pm$0.09 & 86.53$\pm$0.07 \\
 & F-SAM$_{(300)}$ & 97.84$\pm$0.05 & 86.70$\pm$0.14 \\
 & SGD-SALR & / & 85.69$\pm$0.10 \\
 & LookSAM & 97.22$\pm$0.05 & 83.76$\pm$0.45 \\
 & ESAM & 97.81$\pm$0.01 & 85.56$\pm$0.05 \\
 & SAF & 97.34$\pm$0.06 & 84.71$\pm$0.01 \\
 & BSAM & \textbf{97.96$\pm$0.10} & 86.20$\pm$0.06 \\
 & GEAR-SAM & 97.81$\pm$0.01 & \textbf{86.90$\pm$0.27} \\
\hline
\end{tabular}
\end{table}

\subsection{Top Eigenvalues of Hessian}

The maximum Hessian eigenvalue is commonly viewed as a potential
sharpness/flatness metric
\cite{jiang2019fantastic,jastrzkebski2017three,chaudhari2019entropy}.
To offer a more quantitative assessment of sharpness, we compare the maximum
Hessian eigenvalues obtained from SGD, SAM, and GEAR-SAM. The resulting
spectra provide complementary evidence about both the sharpest local direction
and the overall distribution of dominant curvature.

\begin{figure*}[t]
    \centering
    \begin{minipage}[t]{0.30\textwidth}
        \centering
        \includegraphics[width=\linewidth]{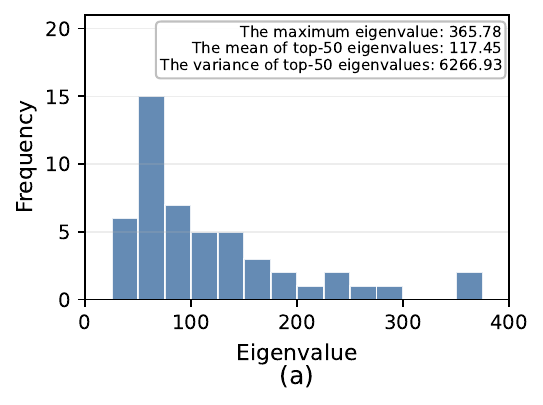}
    \end{minipage}
    \hfill
    \begin{minipage}[t]{0.30\textwidth}
        \centering
        \includegraphics[width=\linewidth]{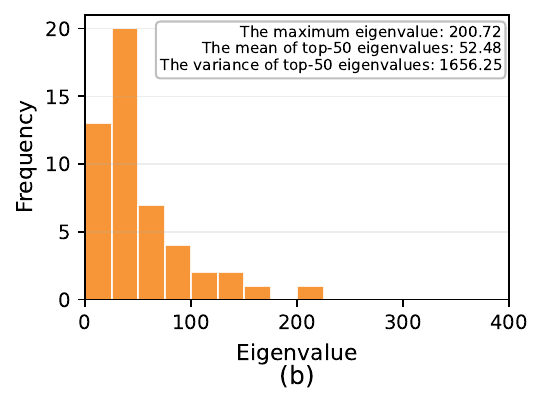}
    \end{minipage}
    \hfill
    \begin{minipage}[t]{0.30\textwidth}
        \centering
        \includegraphics[width=\linewidth]{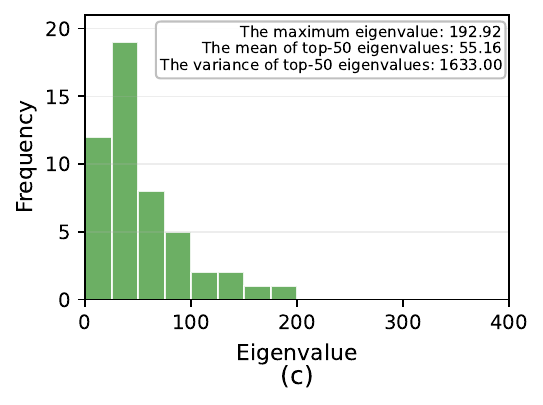}
    \end{minipage}
    \caption{Histograms of the top-50 Hessian eigenvalues of ResNet-18 trained
    on CIFAR-100 with (a) SGD, (b) SAM, and (c) GEAR-SAM. Each panel also
    reports the maximum eigenvalue and the mean and variance of the top-50
    eigenvalues.}
    \label{fig:hessian_top50}
\end{figure*}

We analyze the Hessian spectra of ResNet-18 trained on CIFAR-100 with SGD,
SAM, and GEAR-SAM using the CIFAR-100 test set. We compute the Hessian
spectrum using PyHessian \cite{yao2020pyhessian}. For each method, we report
the maximum eigenvalue of the Hessian, a histogram of the top-50 Hessian
eigenvalues, and the mean and variance of these top-50 Hessian eigenvalues.
As shown in Fig.~\ref{fig:hessian_top50}, both SAM and GEAR-SAM
substantially shift the dominant Hessian spectrum toward smaller values
compared with SGD. GEAR-SAM obtains the smallest maximum eigenvalue, reducing
it from 365.78 for SGD and 200.72 for SAM to 192.92. Its variance is also
slightly lower than that of SAM (1633.00 versus 1656.25), suggesting that
GEAR-SAM suppresses extreme high-curvature directions without broadening the
dominant spectrum. GEAR-SAM does not outperform SAM on every spectral
statistic: its top-50 mean is 55.16, compared with 52.48 for SAM. This result
indicates that the benefit of block-wise radius allocation is concentrated in
reducing the sharpest local direction rather than uniformly shrinking all
dominant eigenvalues. Overall, the spectrum supports a more precise conclusion:
GEAR-SAM reaches a curvature profile comparable to SAM while further reducing
the worst-case local sharpness measured by the maximum Hessian eigenvalue.

\subsection{Robustness to Label Noise}

Since prior studies have shown that SAM is robust to label noise, this
section evaluates the effect of applying GEAR-SAM in the classical noisy-label
setting for CIFAR-10 and CIFAR-100. We evaluate GEAR-SAM under symmetric label
noise by randomly flipping training labels
\cite{huang2019o2u,deng2026bilateral}. The training settings are the same as those in the image classification
experiments, except that the perturbation radii of both SAM and GEAR-SAM are
set to $\rho=0.05$ for all noise rates on both datasets.

As shown in Table~\ref{tab:noise_results}, for CIFAR-10, the accuracy of
models optimized with SGD decreases rapidly as the noise rate increases,
dropping to only 28.91\% when the noise rate reaches 80\%. In contrast, the
accuracy of models optimized with SAM and GEAR-SAM decreases more slowly as
the label noise increases, and both methods maintain over 70\% accuracy at the
80\% noise rate. GEAR-SAM achieves the best results at all four noise rates,
reaching 92.90\%, 90.91\%, 87.21\%, and 79.15\%, respectively.

For CIFAR-100, we observe a similar overall trend. At the 20\% noise rate,
SAM obtains slightly higher accuracy than GEAR-SAM. When the noise rate rises
to 40\%, 60\%, and 80\%, however, GEAR-SAM achieves 57.86\%, 50.45\%, and
35.96\%, respectively, consistently outperforming both SGD and SAM. In
particular, when the noise rate reaches 80\%, SAM falls below SGD, whereas
GEAR-SAM still maintains substantially higher accuracy. These observations
indicate that GEAR-SAM can converge effectively in the presence of extensive
incorrect supervision, facilitates learning from the remaining clean labels,
and mitigates overfitting to noisy labels. Overall, GEAR-SAM improves the
robustness of sharpness-aware training and provides stronger generalization
under moderate and severe label noise.

\begin{table}[t]
\centering
\scriptsize
\caption{Test accuracy (\%) of ResNet-18 under symmetric label noise.}
\label{tab:noise_results}
\setlength{\tabcolsep}{4pt}
\begin{tabular}{llcccc}
\hline
Dataset & Method & 20\% & 40\% & 60\% & 80\% \\
\hline
\multirow{3}{*}{CIFAR-10} & SGD & 85.94 & 70.25 & 48.50 & 28.91 \\
 & SAM & 90.44 & 79.71 & 67.22 & 76.94 \\
 & GEAR-SAM & \textbf{92.90} & \textbf{90.91} & \textbf{87.21} & \textbf{79.15} \\
\hline
\multirow{3}{*}{CIFAR-100} & SGD & 65.42 & 48.91 & 31.46 & 12.32 \\
 & SAM & \textbf{69.10} & 55.06 & 35.59 & 9.82 \\
 & GEAR-SAM & 68.11 & \textbf{57.86} & \textbf{50.45} & \textbf{35.96} \\
\hline
\end{tabular}
\end{table}

\subsection{Transfer Learning}

Transfer learning leverages a model trained on one task to improve
performance on a related task. By fine-tuning a pretrained model or using its
learned representations, transfer learning can reduce the training time and
data requirement of the target task, and often improves performance when the
target dataset is limited. Previous studies have demonstrated the effectiveness
of SAM and its variants in transfer learning \cite{foret2020sharpness,li2024friendly}.
In this section, we evaluate the performance of GEAR-SAM in transfer learning
tasks.

Specifically, we apply SGD, SAM, and GEAR-SAM to fine-tune
EfficientNet-B0 and ResNet-50 pretrained on ImageNet. Weights are initialized
from publicly available checkpoints, except for the final classification
layer, which is resized according to the number of target classes and randomly
initialized. We train the models for 30 epochs with a batch size of 128. The
initial learning rate is set to 0.01 with cosine learning rate decay. Weight
decay is set to $1\times10^{-5}$ for EfficientNet-B0 and $1\times10^{-4}$ for
ResNet-50. For SAM and GEAR-SAM, we use SGD as the base optimizer. The
perturbation radius of SAM is set to 0.05. For GEAR-SAM, the perturbation
radius $\rho$ is set to 0.2, and the EMA factor $\beta$ is set to 0.9. We do
not use any data augmentation for Flowers102, Stanford Cars, and Oxford-IIIT
Pet. For CIFAR-10 and CIFAR-100, we employ the same data augmentations as the
image classification experiments.

As shown in Table~\ref{tab:transfer_results}, SAM exhibits stronger
generalization than SGD on both EfficientNet-B0 and ResNet-50, and GEAR-SAM
further improves SAM in most transfer-learning tasks. On EfficientNet-B0,
GEAR-SAM achieves the best results on all five datasets, with particularly
large gains on Flowers102, Stanford Cars, and Oxford-IIIT Pet. On ResNet-50,
GEAR-SAM obtains the best accuracy on CIFAR-100, Flowers102, and Stanford
Cars, while remaining competitive on CIFAR-10 and Oxford-IIIT Pet. These
results indicate that the proposed block-wise radius allocation is effective
not only when training from scratch, but also when adapting pretrained
representations to downstream tasks.

\begin{table}[t]
\centering
\scriptsize
\caption{Test accuracy (\%) on transfer-learning benchmarks.}
\label{tab:transfer_results}
\setlength{\tabcolsep}{2.5pt}
\begin{tabular}{llccc}
\hline
Backbone & Dataset & SGD & SAM & GEAR-SAM \\
\hline
\multirow{5}{*}{EfficientNet-B0} & CIFAR-10 & 97.32$\pm$0.07 & 97.48$\pm$0.07 & \textbf{97.95$\pm$0.03} \\
 & CIFAR-100 & 87.05$\pm$0.11 & 87.27$\pm$0.21 & \textbf{88.28$\pm$0.05} \\
 & Flowers102 & 77.11$\pm$0.36 & 78.04$\pm$0.37 & \textbf{84.04$\pm$0.24} \\
 & Stanford Cars & 74.47$\pm$0.55 & 75.75$\pm$0.41 & \textbf{80.54$\pm$0.45} \\
 & Oxford-IIIT Pet & 87.89$\pm$0.43 & 88.30$\pm$0.16 & \textbf{91.91$\pm$0.21} \\
\hline
\multirow{5}{*}{ResNet-50} & CIFAR-10 & 97.28$\pm$0.05 & \textbf{97.78$\pm$0.02} & 97.77$\pm$0.04 \\
 & CIFAR-100 & 86.20$\pm$0.18 & 87.48$\pm$0.04 & \textbf{87.80$\pm$0.04} \\
 & Flowers102 & 86.74$\pm$0.16 & 87.53$\pm$0.13 & \textbf{89.29$\pm$0.08} \\
 & Stanford Cars & 80.02$\pm$0.24 & 83.69$\pm$0.19 & \textbf{86.69$\pm$0.10} \\
 & Oxford-IIIT Pet & 93.11$\pm$0.17 & \textbf{93.14$\pm$0.14} & 92.96$\pm$0.11 \\
\hline
\end{tabular}
\end{table}

Additional results comparing alternative block partitions are provided in the
appendix.

\subsection{Parameter Studies}

We examine the sensitivity of GEAR-SAM to its main hyperparameters. The
perturbation radius $\rho$ controls the size of the neighborhood, while the
EMA factor $\beta$ controls the stability of the block sensitivity estimate.
Tables~\ref{tab:rho_study} and~\ref{tab:beta_study} show that GEAR-SAM performs
best around $\rho=0.2$ and $\beta=0.9$ on ResNet-18 with CIFAR-100. Smaller
radii provide weaker sharpness regularization, while larger radii may harm
optimization. Smaller $\beta$ values make the allocation close to
instantaneous gradients, while larger values adapt too slowly.

\begin{table}[t]
\centering
\scriptsize
\caption{Effect of perturbation radius $\rho$ on CIFAR-100 with ResNet-18.}
\label{tab:rho_study}
\begin{tabular}{lccccc}
\hline
\(\rho\) & 0.10 & 0.15 & 0.20 & 0.25 & 0.30 \\
\hline
Accuracy & 81.45 & 81.52 & \textbf{81.83} & 81.18 & 81.22 \\
\hline
\end{tabular}
\end{table}

\begin{table}[t]
\centering
\scriptsize
\caption{Effect of EMA factor $\beta$ on CIFAR-100 with ResNet-18.}
\label{tab:beta_study}
\begin{tabular}{lccccc}
\hline
\(\beta\) & 0.80 & 0.85 & 0.90 & 0.95 & 0.99 \\
\hline
Accuracy & 81.55 & 81.56 & \textbf{81.83} & 81.38 & 81.27 \\
\hline
\end{tabular}
\end{table}

\FloatBarrier
\section{Conclusion}
In this paper, we revisited SAM from a curvature-aware block-wise allocation perspective. We showed that the practical SAM perturbation is optimal for the first-order inner maximization, but the second-order sharpness objective indicates that curvature information should also guide perturbations. We further showed that standard SAM already performs an implicit block-wise radius allocation, where the radius of each block depends only on instantaneous minibatch gradient norms. Based on these observations, we proposed GEAR-SAM, which estimates block sensitivity using EMA-based gradient energy and allocates the fixed SAM radius through a closed-form constrained alignment rule. GEAR-SAM preserves the global SAM budget and requires no explicit Hessian or Fisher computation. Experiments on image classification, transfer learning, noisy-label learning, and partition and parameter studies demonstrate that the proposed allocation strategy improves the effectiveness and robustness of SAM-based optimization.

\bibliographystyle{IEEEtran}
\bibliography{references}

\clearpage
\section*{Appendix}

This appendix provides additional details supporting the design of GEAR-SAM.
It gives complete derivations for the trust-region motivation, the stochastic
gradient second moment, the EMA estimator, and the budget-preserving
allocation rule. These results establish exact properties of the proposed
surrogate while keeping separate the assumptions used to motivate its
curvature-related sensitivity signal.

\subsection*{Trust-region Conditions and SAM Decomposition}

For the quadratic approximation, define
\begin{equation}
    Q_t(\boldsymbol{\epsilon})
    =
    \mathbf{g}_t^{\top}\boldsymbol{\epsilon}
    +
    \frac{1}{2}\boldsymbol{\epsilon}^{\top}
    \mathbf{H}_t\boldsymbol{\epsilon}.
\end{equation}
The Lagrangian for maximizing $Q_t$ subject to
$\|\boldsymbol{\epsilon}\|_2^2\le\rho^2$ is
\begin{equation}
    \mathcal{L}(\boldsymbol{\epsilon},\lambda)
    =
    Q_t(\boldsymbol{\epsilon})
    -
    \lambda(\|\boldsymbol{\epsilon}\|_2^2-\rho^2),
    \quad \lambda\ge 0.
\end{equation}
By the classical trust-region optimality theorem
\cite{more1983computing}, a global maximizer
$\boldsymbol{\epsilon}_t^\star$ admits a multiplier $\lambda_t$ satisfying
\begin{equation}
\begin{aligned}
    \mathbf{g}_t+\mathbf{H}_t\boldsymbol{\epsilon}_t^\star
    &=2\lambda_t\boldsymbol{\epsilon}_t^\star,\\
    \|\boldsymbol{\epsilon}_t^\star\|_2^2&\le\rho^2,\\
    \lambda_t(\|\boldsymbol{\epsilon}_t^\star\|_2^2-\rho^2)&=0,\\
    2\lambda_t\mathbf{I}-\mathbf{H}_t&\succeq0.
\end{aligned}
\label{eq:append_trust_conditions}
\end{equation}
These conditions are also sufficient. To see this, let
$\mathbf{d}=\boldsymbol{\epsilon}-\boldsymbol{\epsilon}_t^\star$ for any
feasible $\boldsymbol{\epsilon}$. Stationarity gives
\begin{equation}
\begin{aligned}
    Q_t(\boldsymbol{\epsilon})-Q_t(\boldsymbol{\epsilon}_t^\star)
    &=2\lambda_t(\boldsymbol{\epsilon}_t^\star)^{\top}\mathbf{d}
      +\frac{1}{2}\mathbf{d}^{\top}\mathbf{H}_t\mathbf{d}\\
    &\le -\frac{1}{2}\mathbf{d}^{\top}
      (2\lambda_t\mathbf{I}-\mathbf{H}_t)\mathbf{d}
    \le 0,
\end{aligned}
\end{equation}
where the first inequality follows from feasibility and complementary
slackness. If $2\lambda_t\mathbf{I}-\mathbf{H}_t\succ0$, stationarity yields
Eq.~\eqref{eq:second_order_direction}. If it is singular, a null-space
component may be required; this is the classical hard case. GEAR-SAM does not
attempt to solve either case online.

Under the block-diagonal interpretation in the main text, let
$a_{b,t}=\|\mathbf{g}_{b,t}\|_2$ and
$q_{b,t}=\mathbf{u}_{b,t}^{\top}\mathbf{H}_{bb,t}\mathbf{u}_{b,t}$. For an
active block coordinate with $r_{b,t}>0$, the KKT stationarity condition
becomes
\begin{equation}
    a_{b,t}+q_{b,t}r_{b,t}=2\lambda_t r_{b,t}.
\end{equation}
Whenever $2\lambda_t>\max_b q_{b,t}$, this gives
\begin{equation}
    r_{b,t}
    =
    \frac{a_{b,t}}{2\lambda_t-q_{b,t}}.
    \label{eq:append_kkt_radius}
\end{equation}
For a fixed feasible multiplier and comparable $a_{b,t}$, the expression
increases with directional curvature $q_{b,t}$. It motivates using
curvature-related block information but does not derive GEAR-SAM's surrogate
allocation rule.

For completeness, when $\mathbf{g}_t\neq\mathbf{0}$, Observation~1 follows
directly from the orthogonal parameter partition:
\begin{equation}
    \|\mathbf{g}_t\|_2^2
    =
    \sum_{b=1}^{B}\|\mathbf{g}_{b,t}\|_2^2.
\end{equation}
Substituting Eq.~\eqref{eq:sam_block_radius} gives
\begin{equation}
    \sum_{b=1}^{B}(r^{\mathrm{SAM}}_{b,t})^2
    =
    \rho^2
    \frac{\sum_b\|\mathbf{g}_{b,t}\|_2^2}
    {\sum_j\|\mathbf{g}_{j,t}\|_2^2}
    =\rho^2.
\end{equation}
If $\mathbf{g}_t=\mathbf{0}$, SAM uses the zero perturbation by convention,
and the block-wise radii are all set to zero.

\subsection*{Minibatch Gradient Second Moment}

Condition on $\mathbf{w}_t$ and let
$\mathbf{z}_{b,i}=\nabla_{\mathbf{w}_b}\ell(\mathbf{w}_t;x_i)$ be IID
per-example gradients with mean $\boldsymbol{\mu}_{b,t}$ and second-moment
block $\mathbf{C}_{bb,t}$. For a minibatch mean
$\mathbf{g}_{b,t}=m^{-1}\sum_{i=1}^{m}\mathbf{z}_{b,i}$,
\begin{equation}
\begin{aligned}
    \mathbb{E}\|\mathbf{g}_{b,t}\|_2^2
    &=\frac{1}{m^2}
      \mathbb{E}\left\|\sum_{i=1}^{m}\mathbf{z}_{b,i}\right\|_2^2\\
    &=\frac{1}{m^2}\left(
      m\,\mathbb{E}\|\mathbf{z}_{b,1}\|_2^2
      +m(m-1)\|\boldsymbol{\mu}_{b,t}\|_2^2\right)\\
    &=\frac{1}{m}\operatorname{Tr}(\mathbf{C}_{bb,t})
      +\left(1-\frac{1}{m}\right)
       \|\boldsymbol{\mu}_{b,t}\|_2^2.
\end{aligned}
\end{equation}
This proves Eq.~\eqref{eq:minibatch_second_moment}. For negative
log-likelihoods, $\mathbf{C}_t$ is the Fisher only when its expectation is
taken under the model distribution. Under the empirical data distribution it
is a gradient second moment, and neither it nor the minibatch quantity above
is generally equal to the Hessian. Near stationarity, the mean-gradient term
is small and the second-moment trace dominates up to the common factor $1/m$.

\subsection*{EMA Expectation and Variance}

Unrolling Eq.~\eqref{eq:ema_energy} with $s_{b,0}=0$ gives
\begin{equation}
    s_{b,t}
    =
    (1-\beta)\sum_{k=0}^{t-1}\beta^k e_{b,t-k}.
\end{equation}
Under local stationarity, $\mathbb{E}[e_{b,t-k}]=\bar e_b$ throughout this
window. Therefore,
\begin{equation}
    \mathbb{E}[s_{b,t}]
    =
    (1-\beta)\bar e_b\sum_{k=0}^{t-1}\beta^k
    =
    (1-\beta^t)\bar e_b,
\end{equation}
which proves Eq.~\eqref{eq:ema_expectation}. If the centered stochastic
energies are additionally independent over iterations with variance
$\sigma_b^2$, then
\begin{equation}
\begin{aligned}
    \operatorname{Var}(s_{b,t})
    &=(1-\beta)^2\sigma_b^2
      \sum_{k=0}^{t-1}\beta^{2k}\\
    &=\frac{1-\beta}{1+\beta}
      (1-\beta^{2t})\sigma_b^2.
\end{aligned}
\label{eq:append_ema_variance}
\end{equation}
Thus, under this idealized independent-noise model, the asymptotic variance is
reduced by $(1-\beta)/(1+\beta)$ relative to an instantaneous energy. The
formula is an estimator property under the stated assumptions, not a claim
that optimization gradients are exactly stationary or independent.

\subsection*{Closed-form Allocation and Budget Preservation}

Let $\mathbf{s}_t\neq\mathbf{0}$ be nonnegative. For every feasible radius
vector $\mathbf{r}_t$, Cauchy--Schwarz gives
\begin{equation}
    \mathbf{s}_t^{\top}\mathbf{r}_t
    \le
    \|\mathbf{s}_t\|_2\|\mathbf{r}_t\|_2
    \le
    \rho\|\mathbf{s}_t\|_2.
\end{equation}
Equality in both inequalities is attained uniquely at
\begin{equation}
    \mathbf{r}_t^\star
    =
    \rho\frac{\mathbf{s}_t}{\|\mathbf{s}_t\|_2},
\end{equation}
which is nonnegative and proves Proposition~1. It also gives
\begin{equation}
    \sum_{b=1}^{B}(r_{b,t}^{\star})^2=\rho^2.
\end{equation}
For exact unit block directions, the concatenated perturbation therefore has
norm $\rho$. With the stabilized directions used in Algorithm~\ref{alg:gear_sam},
each direction has norm at most one, so the realized perturbation instead
satisfies $\|\boldsymbol{\epsilon}_t\|_2\le\rho$. A common positive rescaling
of all sensitivity scores leaves $\mathbf{r}_t^\star$ unchanged.
If $\mathbf{s}_t=\mathbf{0}$, the alignment objective is identically zero
over the feasible set. In this degenerate case, Algorithm~\ref{alg:gear_sam}
sets all radii to zero, which is feasible and avoids injecting an arbitrary
perturbation before any block-sensitivity signal is available.

\subsection*{Coarse Block Grouping}

Let a block contain $K$ tensors with energy estimates
$e_{1,t},\ldots,e_{K,t}$. Suppose each tensor-level estimate can be decomposed
as $e_{i,t}=\mu_i+\xi_{i,t}$, where $\xi_{i,t}$ is zero-mean minibatch noise.
Tensor-wise grouping assigns a separate radius to every noisy estimate. In
contrast, coarse block grouping uses the aggregate
\begin{equation}
    e_{b,t}
    =
    \sum_{i=1}^{K}e_{i,t}
    =
    \sum_{i=1}^{K}\mu_i+\sum_{i=1}^{K}\xi_{i,t}.
\end{equation}
If the noise terms are weakly correlated, the relative fluctuation of the
aggregate score is smaller than that of individual tensor scores. For example,
with independent noise of variance $\sigma^2$ and comparable means $\mu$, the
coefficient of variation changes from $\sigma/\mu$ for a tensor-level score
to $\sigma/(\sqrt{K}\mu)$ for the aggregated block score. Coarse grouping
therefore improves the stability of online allocation.

Coarse grouping also preserves the joint perturbation direction of tensors
that operate at the same representation scale. Tensor-wise grouping can
normalize each tensor independently and fragment a functional block into many
small allocation decisions. Coarse block grouping avoids this fragmentation
while still distinguishing parts of the network with different spatial
resolution, channel width, and optimization roles. This provides the
theoretical rationale for the architecture-aware partition used in GEAR-SAM.

\subsection*{Structured Partitioning Analysis}

We further investigate alternative block partitions. Tensor-wise grouping
assigns an independent sensitivity score and radius to every weight or bias
tensor. Our default coarse block-wise grouping assigns one score to each
architecture-level functional block, such as \texttt{layer1} in ResNet-18.
Fine block-wise grouping assigns a separate score to every residual block,
while keeping the stem and classifier as separate blocks.

As shown in Table~\ref{tab:partition_strategies}, all structured partitions
clearly outperform standard SAM. The default coarse block-wise GEAR-SAM
improves SAM from 81.08 to 81.83 on ResNet-18 and from 84.71 to 85.49 on
WideResNet-28-10. Fine block-wise grouping performs slightly better still,
reaching 81.84 and 85.60, respectively. These results show that the benefit
of GEAR-SAM is not tied to a single manually selected partition. More broadly,
preserving functional units while avoiding excessively small groups appears
important for reliable online allocation. Tensor-wise grouping also improves
SAM, but its larger number of independent allocation variables can introduce
noisier sensitivity estimates and excessive allocation freedom. The small
advantage of fine blocks suggests that the most effective granularity can
depend on the architecture; learning an adaptive block partition is an
interesting direction for future work.

\begin{table}[t]
\centering
\scriptsize
\caption{Comparison of partition strategies on CIFAR-100.}
\label{tab:partition_strategies}
\setlength{\tabcolsep}{3pt}
\begin{tabular}{llcc}
\hline
Method & Grouping & ResNet-18 & WRN-28-10 \\
\hline
SAM & -- & 81.08 & 84.71 \\
\multirow{3}{*}{GEAR-SAM} & Tensor-wise & 81.67 & 85.24 \\
 & Coarse block-wise & 81.83 & 85.49 \\
 & Fine block-wise & \textbf{81.84} & \textbf{85.60} \\
\hline
\end{tabular}
\end{table}

\subsection*{Full Block-Wise Perturbation Dynamics}

\begin{figure*}[t]
    \centering
    \includegraphics[width=\textwidth]{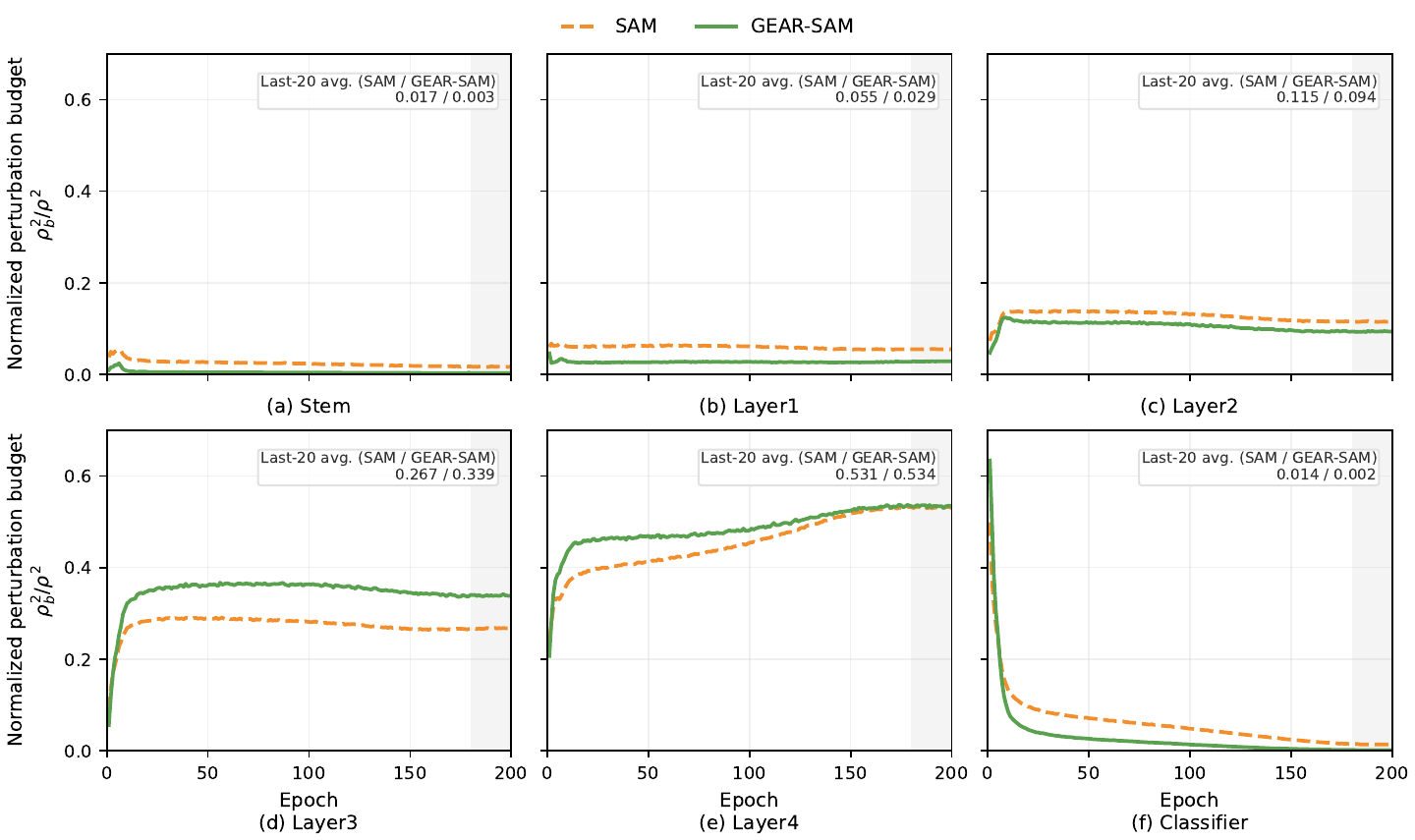}
    \caption{Full block-wise perturbation dynamics of SAM and GEAR-SAM on
    ResNet-18 trained on CIFAR-100. Each panel shows the normalized
    perturbation-budget share $\rho_{b,t}^{2}/\rho^{2}$ of one network block.
    The gray region denotes the final 20 epochs, and each annotation reports
    the corresponding SAM and GEAR-SAM averages. All panels use the same
    vertical scale.}
    \label{fig:append_block_dynamics}
\end{figure*}

For completeness, Fig.~\ref{fig:append_block_dynamics} reports the normalized
budget shares of all six coarse blocks over the full 200-epoch training run.
The trajectories provide the long-horizon context for the early redistribution
shown in Fig.~\ref{fig:block_dynamics}: the allocation is not a uniform
rescaling of SAM, but a redistribution of the same global perturbation budget.

\FloatBarrier


\end{document}